\documentclass[conference]{IEEEtran}

\usepackage{soul}
\usepackage{multirow}
\usepackage{url}
\usepackage[utf8]{inputenc}
\usepackage{graphicx}
\usepackage{amsmath}
\usepackage{amsthm}
\usepackage{booktabs}
\usepackage{algorithm}
\usepackage{algorithmic}
\usepackage{hyperref}
\usepackage{url}
\usepackage{graphicx}
\usepackage{amsmath}
\usepackage{bbm}

\newcommand{\mb}{\mathbf}

\newcommand{\mc}{\mathcal}

\newtheorem{definition}{\textsc{Definition}}

\newcommand{\our}{\textsc{G5}}
\newcommand{\gbert}{\textsc{Graph-Bert}}

\newcommand{\transformer}{\textsc{Transformer}}
\newcommand{\bert}{\textsc{Bert}}

\newcommand{\gcn}{\textsc{GCN}}
\newcommand{\gat}{\textsc{GAT}}
\newcommand{\loopy}{\textsc{LoopyNet}}
\begin{document}

\title{{\our}: A Universal {\gbert} for Graph-to-Graph Transfer and Apocalypse Learning}



\author{\IEEEauthorblockN{Jiawei Zhang}
\IEEEauthorblockA{IFM Lab, Florida State University, Tallahassee, FL, USA}
jiawei@ifmlab.org}

\maketitle

\begin{abstract}

The recent {\gbert} model introduces a new approach to learning graph representations merely based on the attention mechanism. {\gbert} provides an opportunity for transferring pre-trained models and learned graph representations across different tasks within the same graph dataset. In this paper, we will further investigate the \underline{g}raph-to-\underline{g}raph transfer of a universal \textsc{\underline{G}raph-Bert} for \underline{g}raph representation learning across different \underline{g}raph datasets, and our proposed model is also referred to as the ``{\our}'' for simplicity. Many challenges exist in learning {\our} to adapt the distinct input and output configurations for each graph data source, as well as the information distributions differences. {\our} introduces a pluggable model architecture: (a) each data source will be pre-processed with a unique input representation learning component; (b) each output application task will also have a specific functional component; and (c) all such diverse input and output components will all be conjuncted with a universal {\gbert} core component via an \textit{input size unification layer} and an \textit{output representation fusion layer}, respectively. 

The {\our} model removes the last obstacle for cross-graph representation learning and transfer. For the graph sources with very sparse training data, the {\our} model pre-trained on other graphs can still be utilized for representation learning with necessary fine-tuning. What's more, the architecture of {\our} also allows us to learn a supervised functional classifier for data sources without any training data at all. Such a problem is also named as the \textit{Apocalypse Learning} task in this paper. Two different label reasoning strategies, i.e., Cross-Source Classification Consistency Maximization (CCCM) and Cross-Source Dynamic Routing (CDR), are introduced in this paper to address the problem. The preliminary experimental results on several benchmark graph datasets can demonstrate the effectiveness of {\our} on graph-to-graph transfer and representation learning.

\end{abstract}


\section{Introduction}\label{sec:introduction}

A brand new graph neural network named {\gbert} (Graph based {\bert}) is introduced in \cite{zhang2020graph} for graph data representation learning. Different from conventional graph neural networks \cite{Kipf_Semi_CORR_16,Velickovic_Graph_ICLR_18,Li_Deeper_CORR_18,sun2019adagcn,DBLP:journals/corr/abs-1907-02586}, via linkless subgraph batching, {\gbert} redefines the conventional graph representation learning problem as the target node instance representation learning within individual learning context instead. One of the great advantages of such a new learning setting is that {\gbert} can effectively get rid of many common learning effectiveness and efficiency problems, e.g., suspended animation \cite{Zhang2019GResNetGR} and hard to parallelize, with the existing graph neural networks. Also it enables the pre-training and fine-tuning of {\gbert} across different learning tasks on the same graph dataset, which has transformative impacts on building functional model pipelines for graph learning.

In this paper, we will further explore the transfer of {\gbert} across different graph datasets, which still remains a great challenge and an open problem by this context so far. To be more precise, we propose to learn {\gbert} with multiple different graph datasets, which have totally different properties, e.g., graph sizes, graph structures, input feature space and output label space. What's more, the learned {\gbert} on one or several source graph dataset(s) can be further transferred as the pre-trained model for other graph dataset(s) suffering from the lack of training data. For each of these graph data datasets, multiple different application tasks can also be studied concurrently, which may or may not have correlations with each other.

To address such a problem, a novel learning model, i.e., {\our}, will be introduced in this paper, where the five Gs correspond to the ``\underline{g}raph-to-\underline{g}raph transfer of a universal \textsc{\underline{G}raph-Bert} for \underline{g}raph representation learning across different \underline{g}raph datasets''. {\our} effectively extends the {\gbert} model for the cross-graph representation learning, which brings about lots of new challenges and new opportunities at the same time.

On the one hand, to learn the {\our} model, we may need to explore many great challenges in handling the different graph property differences and the different objectives of diverse learning tasks. To be more specific, {\our} introduces a pluggable model architecture: (a) each data source will be pre-learned with a unique input component for data pre-processing; (b) each output application task will also have a specific functional component for computing the output; and (c) all such diverse input and output components will be conjuncted with a universal {\gbert} core component in {\our} via an \textit{input size unification layer} \cite{zhang2020segmented} and an \textit{output representation fusion layer} \cite{zhang2020graph}, respectively. 

On the other hand, in addition to building the functional model pipelines across graphs for representation learning, a successfully learned {\our} will also allow us to explore some new yet challenging problems. Besides the model transfer to graph sources with limited training data, the architecture of {\our} also allows us to learn a supervised functional classifier for certain graph sources without any training data at all, which is also named as the \textit{Apocalypse Learning} (AL) problem formally in this paper. It should be easy to identify that the \textit{apocalypse learning} task is different from the well-studied \textit{zero-shot learning} task \cite{NIPS2013_5027}. Here, we would like to further clearly illustrate their differences: (1) \textit{apocalypse learning} is for multi-dataset but \textit{zero-shot learning} focuses on one dataset; (2) \textit{apocalypse learning} uses no training data in the target data source but \textit{zero-shot learning} uses training data; and (3) \textit{apocalypse learning} requires no prior knowledge but \textit{zero-shot learning} usually needs to know prior class representations or correlations in advance.

We summarize our contributions in this paper as follows:
\begin{itemize}

\item \textbf{A Universal GNN}: We introduce a new graph neural network model in this paper for multi-graph concurrent representation learning. To adapt the diverse input and output configuration distinctions, as well as the graph information distributions differences, {\our} introduces a pluggable model architecture which can be tied up with many different input and output components. All such diverse input and output components will be conjuncted with a universal {\gbert} core component in {\our} via the \textit{input size unification layer} and \textit{output representation fusion layer}, respectively. 

\item \textbf{Pre-Train \& Transfer \& Fine-Tune}: To learn various application task objectives, {\our} will be pre-trained on multiple graphs in a hybrid manner with multiple different learning tasks, which also define the output component pool involving various supervised and unsupervised learning tasks. Meanwhile, a pre-trained {\our} can also be transferred and applied to new graph data sources either directly or with necessary fine-tuning in a similarly hybrid manner. There is no specific correlation requirements on these fine-tuning tasks, which can be totally different from those pre-training tasks on the source graph(s) actually.

\item \textbf{Apocalypse Learning}: Besides investigating the model transfer to graph sources with limited training data, in this paper, we also introduce a new learning problem, i.e., \textit{apocalypse learning}, which aims to build a classifier on certain target graph source without any training data at all. Based on the learning results of the hybrid tasks on other graph datasets, {\our} introduces two different strategies, i.e., Cross-Source Classification Consistency Maximization (CCCM) and Cross-Source Dynamic Routing (CDR), to reason for the labels in the target graph source in this paper.

\end{itemize}

The remaining parts of this paper are organized as follows. Definitions of several important terminologies and the formulation of the studied problem will be provided in Section~\ref{sec:formulate}. Detailed information about the {\our} model will be introduced in Section~\ref{sec:method}, and the two reasoning strategies to address the {apocalypse learning} problem will be discussed in Section~\ref{sec:analysis}. The effectiveness of {\our} will be tested in Section~\ref{sec:experiment}. Finally, we will introduce the related work in Section~\ref{sec:related_work} and conclude this paper in Section~\ref{sec:conclusion}.

\section{Notations, Terminology Definition and Problem Formulation}\label{sec:formulate}

In this section, we will first introduce the notations used in this paper. After that, we will provide the definitions of several important terminologies and the studied problem.

\subsection{Notations}\label{subsec:notation}

In the sequel of this paper, we will use the lower case letters (e.g., $x$) to represent scalars or mappings, lower case bold letters (e.g., $\mathbf{x}$) to denote column vectors, bold-face upper case letters (e.g., $\mathbf{X}$) to denote matrices, and upper case calligraphic letters (e.g., $\mathcal{X}$) to denote sets or high-order tensors. Given a matrix $\mathbf{X}$, we denote $\mathbf{X}(i,:)$ and $\mathbf{X}(:,j)$ as its $i_{th}$ row and $j_{th}$ column, respectively. The ($i_{th}$, $j_{th}$) entry of matrix $\mathbf{X}$ can be denoted as either $\mathbf{X}(i,j)$. We use $\mathbf{X}^\top$ and $\mathbf{x}^\top$ to represent the transpose of matrix $\mathbf{X}$ and vector $\mathbf{x}$. For vector $\mathbf{x}$, we represent its $L_p$-norm as $\left\| \mathbf{x} \right\|_p = (\sum_i |\mathbf{x}(i)|^p)^{\frac{1}{p}}$. The Frobenius-norm of matrix $\mathbf{X}$ is represented as $\left\| \mathbf{X} \right\|_F = (\sum_{i,j} |\mathbf{X}(i,j)|^2)^{\frac{1}{2}}$. The element-wise product of vectors $\mathbf{x}$ and $\mathbf{y}$ of the same dimension is represented as $\mathbf{x} \otimes \mathbf{y}$, whose concatenation is represented as $\mathbf{x} \sqcup \mathbf{y}$.

\subsection{Terminology Definitions}\label{subsec:terminology_definition}

Several terminologies will be used in this paper to present the proposed method, which include \textit{graph}, \textit{multi-source graph set} and \textit{linkless subgraph}.

\begin{definition}
(Graph): Formally, we can represent the studied graph data as $G = (\mc{V}, \mc{E}, w, x, y)$, where $\mc{V}$ and $\mc{E}$ denote the sets of nodes and links, respectively. Mapping $w: \mc{E} \to \mathbbm{R}$ projects links to their weights; whereas mappings $x: \mc{V} \to \mc{X}$ and $y: \mc{V} \to \mc{Y}$ can project the nodes to their raw features and labels, respectively.
\end{definition}

Given a graph $G$, its size can be represented by the number of involved nodes, i.e., $|\mc{V}|$. Notations $\mc{X}$ and $\mc{Y}$ used in the above definition denote the feature space and label space, respectively. In this paper, they can also be represented as $\mc{X} = \mathbbm{R}^{d_x}$ and $\mc{Y} = \mathbbm{R}^{d_y}$ (with dimensions $d_x$ and $d_y$) for simplicity. For node $v_i$, we can also simplify its raw feature and label vector representations as $\mb{x}_i = x(v_i) \in \mathbbm{R}^{d_x \times 1}$ and $\mb{y}_i = y(v_i) \in \mathbbm{R}^{d_y \times 1}$. In this paper, we are studying the transfer of {\gbert} across multiple graphs, and the studied graphs can be denoted as the multi-source graph set as follows.

\begin{definition}
(Multi-Source Graph Set): Formally, we can represent the set of $n$ different input graphs that we are studying in this paper as $\mc{G} = \left\{G^{(1)}, G^{(2)}, \cdots, G^{(n)} \right\}$, among which some of them may have very limited or even no training data (i.e., labeled nodes). All these $n$ input graphs can have different properties, e.g., graph sizes, graph structures, node feature space and label space.
\end{definition}

Given a node, e.g., $v_i^{(m)} \in \mc{V}^{(m)}$, in graph $G^{(m)} \in \mc{G}$, based on the approach introduced in \cite{zhang2020graph}, we will be able to sample a unique linkless sub-graph for it involving node $v_i$ and its surrounding node context.

\begin{definition}
(Linkless Subgraph): Given an input graph $G^{(m)}$, we can denote the sampled linkless subgraph for each node $v_i^{(m)}$ in the graph as $g_i^{(m)} = (\mc{V}_i^{(m)}, \emptyset)$. Here, the node set $\mc{V}_i^{(m)} = \{v_i^{(m)}\} \cup \Gamma(v_i^{(m)}, k^{(m)})$ covers both $v_i$ and its top $k^{(m)}$ intimate nearby nodes, and the link set is empty. Furthermore, the batch of linkless subgraphs sampled for all the nodes in graph $G^{(m)}$ can be denoted as $\mc{G}^{(m)} = \left\{g_i^{(m)}\right\}_{v_i^{(m)} \in \mc{V}^{(m)}}$.
\end{definition}

Therefore, for all the graphs covered in $\mc{G}$, we can represent their sampled subgraph batches as $\left\{\mc{G}^{(1)}, \mc{G}^{(2)}, \cdots, \mc{G}^{(n)}\right\}$. According to the experimental studies provided in \cite{zhang2020graph}, different graphs may have different optimal parameters to control the sampled subgraph size, e.g., $k^{(m)}$ for graph $G^{(m)}$. Therefore, the subgraphs sampled in batch $\mc{G}^{(l)}$ will usually have different sizes from those in $\mc{G}^{(m)}, \forall l, m \in \{1, 2, \cdots, n\} \land l \neq m$.

\begin{figure*}[t]
    \centering
    \begin{minipage}{1.0\textwidth}
    	\includegraphics[width=\linewidth]{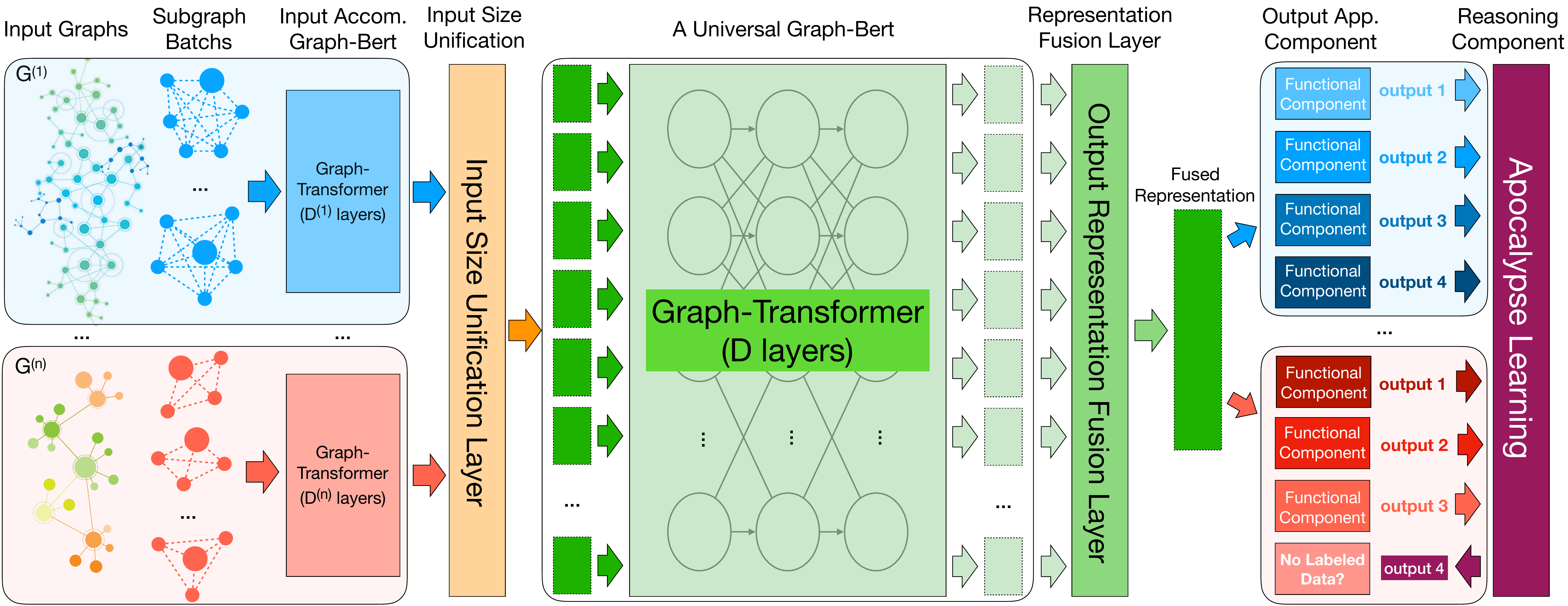}
     \end{minipage}
        \caption{The Architecture of the {\our} Model.}
    	\label{fig:framework}
\end{figure*}

\subsection{Problem Formulation}\label{subsec:formulation}

Based on the above terminology definitions, we can define the problem studied in this paper as follows:

\noindent \textbf{Problem Statement}: Formally, given the multi-source graph set $\mc{G} = \{G^{(1)}, G^{(2)}, \cdots, G^{(n)}\}$ with $n$ different graphs, we aim to learn a shared representation learning mapping $f: \bigcup_{m=1}^n \mc{V}^{(m)} \to \mathbbm{R}^{d_h}$ to learn the representations of nodes in all these $n$ graphs concurrently. Such learned node representations will be further utilized in various downstream application tasks for either pre-training or fine-tuning the model. Furthermore, depending on the learning settings, the mapping pre-trained based on some graphs can also be further transferred to the other graphs with limited even no supervision information directly or with necessary fine-tuning. In this way, it can hopefully help address the labeled data sparsity problem or even the \textit{apocalypse learning} problem for some input graph datasets.


\section{The Proposed Method}\label{sec:method}

In this section, we will introduce the {\our} model architecture in detail. 



\subsection{The Key Challenges}

As introduced in Section~\ref{subsec:terminology_definition}, for the multi-source input graphs $\mc{G}$, a batch of linkless subgraphs can be sampled from them for target node representation learning. To enable the concurrent learning of the {\our} model with all these $n$ graphs in $\mc{G}$, several important differences among these graph datasets cannot be ignored:

\begin{itemize}

\item \textbf{Input Space Difference}: For any two nodes from two different graphs, their raw features can be very different in (1) data types: their feature vectors can be in totally different data types, e.g., image, text, or tags; (2) feature length: the vectors can also have different length; (3) feature domain: for the features of the same type and have the same dimensions, they may also from totally different domains and carry different information, e.g., medical images vs traffic images; and (4) feature distribution: for the identical features in different graph sources, they may follow distinct distributions.

\item \textbf{Model Configuration Difference}: In addition to the input feature space differences aforementioned, there may also exist a lot of model configuration differences in the favored {\gbert} component in {\our} by different graph datasets. For instance, according to \cite{zhang2020graph}, the sampled subgraph size parameter $k$ may affect the learning performance of {\gbert} a lot; whereas different graph datasets may also prefer different parameter $k$s, which may lead to different model configurations.

\item \textbf{Output Space Difference}: Meanwhile, for the downstream application tasks to be studied in {\our} on the same/different graph datasets, they tend to have different output space actually, which may cast certain task oriented requirements on the representation learning process. For instance, the node raw feature reconstruction and graph structure recovery tasks actually focus more on embedding node attributes and graph structures into the learned representations, respectively; whereas the node classification aims at learning a classifer to project nodes to the label space instead.

\end{itemize}

To handle these above differences properly, as illustrated in Figure~\ref{fig:framework}, we design the {\our} model with a pluggable architecture containing several key parts: (1) pluggable input dataset-wise processing components, (2) input size unification interlayer, (3) the universal {\gbert} model shared across graphs, (4) representation fusion interlayer, (5) pluggable task-wise output components for each dataset, and (6) reasoning component for \textit{apocalypse learning}. For each input graph data, it will have a unique input component to handle its initial embeddings based on their unique subgraph batches, which will accommodate the input feature space differences and information distribution differences for {\our}. Meanwhile, each graph dataset will have several output components as multiple pre-train/fine-tune tasks will be studied concurrently, which can handle the output space difference problem. The input size unification interlayer introduced in this paper can effectively accommodate the configures of diverse inputs from different datasets prior to feeding them into the universal {\gbert} model; whereas the representation fusion interlayer will aggregate the learned representations to generate the fused representations for the output components.

The {\our} model will be effectively pre-trained based on the graph datasets with sufficient supervision information, which can be further transferred to the graph datasets lacking enough labeled data with fine-tuning. Furthermore, if certain fine-tuning task on the target graph dataset doesn't contain any supervision information, the \textit{apocalypse learning} based component will be used for label reasoning. In this section, we will introduce the first five components in {\our}, except the reasoning component for \textit{apocalypse learning}, which will be introduced in the next Section~\ref{sec:analysis} in detail.


\subsection{Input Accommodation Component}\label{subsec:input_component}

For presentation simplicity, in this part, we will first ignore the script index for the graphs in the notations. Formally, given the sampled subgraph batch from an input graph $G$, for the target nodes $v_i$ together with their learning context (with $k$ nodes), according to \cite{zhang2020graph}, we can represent its initial embedding vector as
\begin{equation}
\mb{h}_i^{(0)} = \mbox{Aggregate} \left( \mb{e}_i^{x}, \mb{e}_i^{r}, \mb{e}_i^{p}, \mb{e}_i^{d} \right),
\end{equation}
where $\mb{e}_i^{x}$, $\mb{e}_i^{r}$, $\mb{e}_i^{p}$ and $\mb{e}_i^{d} \in \mathbbm{R}^{(k+1)d_e \times 1 }$ denote the embedding vectors based on the raw features, WL based roles, relative positions and the hop based distance as introduced in \cite{zhang2020graph}, respectively. In the notation, $k$ denotes the subgraph sampling parameter, and $d_e$ is the raw embedding feature vector dimension in the graph. The $\mbox{Aggregate}(\cdot)$ function will effectively aggregate the input vectors together, which can be defined in different ways. In this paper, we will follow the previous work, and just define it as the simple vector summation. 

It is easy to know that the raw embedding feature dimension, i.e., $d_e$, in the graph datasets can be different form each other. Also the initial embedding features can lie in different feature spaces for different graph datasets. Therefore, instead of directly feeding vector $\mb{h}_i^{(0)}$ to the universal {\gbert} model, to accommodate the input feature space, {\our} introduces an input component for each graph dataset based on the graph-transformer to project the inputs to a shared feature space of dimension $d_h$ as follows:
\begin{equation}
\begin{cases}
\vspace{8pt}
\mb{H}_i^{(0)} & = \left[\mb{h}_i^{(0)}, \mb{h}_{i,1}^{(0)}, \cdots, \mb{h}_{i,k}^{(0)}\right]^\top ,\\
\mb{H}_i^{(l)} & = \mbox{G-Transformer} \left( \mb{H}_i^{(l-1)}\right), \forall l \in \{1, 2, \cdots, D\},
\end{cases}
\end{equation}
where $D$ denotes the input component depth and the nodes in set $\left\{v_{i,1}, v_{i,2}, \cdots, v_{i,k} \right\} = \Gamma(v_i, k)$ denotes the learning context of $v_i$ in its sampled subgraph. Notation $\mbox{G-Transformer}(\cdot)$ denotes the graph-transformer layers consisting of both the transformer and graph residual terms as introduced in \cite{zhang2020graph}, which will also be defined in the following Equation~(\ref{equ:gtransformer}) in detail. Formally, the finally learned representation matrix $\mb{H}_i^{(D)} \in \mathbbm{R}^{(k+1) \times d_h}$ for the subgraph $g_i$ will be the representation input of the subgraph to the follow-up universal {\gbert} model. 

According to the above descriptions, we can accommodate the input representations for all the sampled subgraphs from all the input graphs, i.e., $\mc{G} =  \left\{G^{(1)}, G^{(2)}, \cdots, G^{(n)} \right\}$. For instance, by adding the graph index superscript into the notations, we can represent such learned nodes' representations from graph $G^{(m)}$ as $\left\{ \mb{H}_i^{(m, D^{(m)})} \right\}_{v_i^{(m)} \in \mc{V}^{(m)}}$, where $\mb{H}_i^{(m, D^{(m)})} \in \mathbbm{R}^{(k^{(m)}+1) \times d_h}$. Here, we may need to add a remark: the input components for the different graphs in $\mc{G}$ will not share the weight parameters, and they can also be in different depths (i.e., $D^{(m)}$) depending on their unique requirements.


\subsection{Input Size Unification Interlayer}

According to the previous subsection, for the input feature dimension, feature domain and distribution differences, they can be effectively handled with the input components consisting of several graph-transformer layers. Meanwhile, it is easy to observe that the accommodated input representations for subgraphs from different graph source still have different configurations, since the subgraph size parameter used in them are usually different, i.e., $k^{(l)} \neq k^{(m)}$ for $G^{(l)}, G^{(m)} \in \mc{G}$. Therefore, prior to feeding them to the universal {\gbert} model, we introduce one more layer to accommodate the input subgraph representation sizes from different graph datasets, which is called the \textit{input size unification interlayer}. There exist different input size unification approaches that can be used, e.g., \textit{full-input strategy}, \textit{padding/pruning strategy} and \textit{segment shifting strategy} as introduced in \cite{zhang2020segmented}. We can take the \textit{padding/pruning strategy} as an example to introduce here, but the other two strategies can be used as well depending on the specific learning settings.

Formally, we can denote the dimension of the inputs for the universal {\gbert} model (to be introduced in the next subsection) as $\mathbbm{R}^{(k+1) \times d_h}$, where the parameter $k$ without superscript denotes the objective subgraph node context size desired by the universal {\gbert} model. Meanwhile, depending on the input subgraph representations and their subgraph size parameters $k^{(m)}$ for graph $G^{(m)}$, the \textit{padding/pruning strategy} based size unification layer will handle them as follows:
\begin{itemize}

\item \textit{Pruning}: If $k^{(m)} > k$, the input has more feature entries than that the universal {\gbert} model can handle. Therefore, the size unification layer will prune the last $k^{(m)} - k$ vector entries from the input, which correspond to the context nodes less relevant to the target node.

\item \textit{No Action}: If $k^{(m)} = k$, the inputs can be handled by the universal {\gbert} directly and no action is necessary to be performed at the size unification layer.

\item \textit{Padding}: If $k^{(m)} < k$, necessary dummy vectors will be needed to be padded to the inputs to increase the involved subgraph node number from $k^{(m)}$ to $k$. We will use the zero padding for simplicity in this paper, which will not dramatically affect the learning results according to \cite{zhang2020segmented} but can introduce more learning time costs.

\end{itemize}
Formally, given the input representation matrix $\mb{H}_i^{(m,D^{(m)})}$ learned for subgraph $g_i^{(m)}$ from graph $G^{(m)}$, we can denote its size-unified output representations as
\begin{equation}
\mb{Z}_i^{(m, 0)} = \mbox{Unify} \left( \mb{H}_i^{(m,D^{(m)})} \right) \in \mathbbm{R}^{(k+1) \times d_h}.
\end{equation}
Similar operators can be applied to all the remaining subgraphs sampled from all these $n$ input graph datasets.


\subsection{Universal {\gbert}}\label{subsec:graph_bert}

The universal {\gbert} model is shared for all the input graph datasets, which can learn the representations based on the inputs iteratively with several layers. Here, we can denote the inputs to {\gbert} from the \textit{input size unification layer} as $\mb{Z}^{(0)} \in \mathbbm{R}^{(k+1) \times d_h}$ without indicating its node index or the graph index in the subscript/superscript. The representation learning component in {\gbert} also contains several layers of the graph-transformers. Formally, at the $l_{th}$ layer, we can represent the learned representation as follows:
\begin{equation}\label{equ:gtransformer}
\begin{aligned}
\hspace{-5pt} &{\mb{Z}}^{(l)} = \mbox{G-Transformer} \left( \mb{Z}^{(l-1)}\right)\\
&= \mbox{softmax} \left(\frac{\mb{Q}^{(l)} (\mb{K}^{(l)})^\top}{\sqrt{d_h}} \right) \mb{V}^{(l)} + \mbox{G-Res} \left( \mb{Z}^{(l-1)}, \mb{X}\right),
\end{aligned}
\end{equation}
where 
\begin{equation}
\begin{cases}
\vspace{3pt}
\mb{Q}^{(l)} & = \mb{Z}^{(l-1)} \mb{W}_Q^{(l)},\\
\vspace{3pt}
\mb{K}^{(l)} & = \mb{Z}^{(l-1)} \mb{W}_K^{(l)},\\
\vspace{3pt}
\mb{V}^{(l)} & = \mb{Z}^{(l-1)} \mb{W}_V^{(l)}.\\
\end{cases}
\end{equation}
In the above equation, $\mb{W}_Q^{(l)}, \mb{W}_K^{(l)}, \mb{W}_K^{(l)} \in \mathbbm{R}^{d_h \times d_h}$ denote the involved variables in the $l_{th}$ layer. In this paper, to simplify the presentation and notations, the hidden representations at different layers in the universal {\gbert} are assumed to have the identical dimension $d_h$ by default. Notation $\mbox{G-Res} \left( \mb{Z}^{(l-1)}, \mb{X}\right)$ defines the graph residual term introduced in \cite{Zhang2019GResNetGR}, and $\mb{X}$ is the raw features of all nodes in the subgraph. For both the shared universal {\gbert} component and the individual graph input components introduced in Section~\ref{subsec:input_component}, we will use the ``\textit{graph-raw}'' residual term in this paper by default. The universal {\gbert} component involved in {\our} will contain $D$ layers, and we can denote the output by the $D_{th}$ layer as $\mb{Z}^{(D)} \in \mathbbm{R}^{(k+1) \times d_h}$.

\subsection{Output Representation Fusion Interlayer}

As illustrated in Figure~\ref{fig:framework}, one more fusion layer is stacked on the universal {\gbert} model to fuse such learned representations to define the ultimate representation vector of the target node, which can be denoted as:
\begin{equation}
{\mb{z}}= \mbox{Fusion} \left( \mb{Z}^{(D)} \right) = \frac{1}{k+1} \sum_{i=1}^{k+1} \mb{Z}^{(D)}(i,:). 
\end{equation}
Many advanced fusion strategies can also be used here, e.g., fusion with further node selections or weighted fusion based on certain attention scores. However, in this paper, we will not explore them and a simple averaging function can be used here to define the above fusion component across all the nodes in the sampled subgraphs. Based on the above descriptions, by bringing the node and graph index subscript/superscript back, we can represent the outputted representations of all the nodes in graph $G^{(m)}$ by the universal {\gbert} component as $\left\{\mb{z}_i^{(m)} \right\}_{v_i^{(m)} \in \mc{V}^{(m)}}$, which will be fed to the following functional components to study various downstream application tasks.


\subsection{Output Application Components}\label{subsec:hybrid_training}

To learn such representations together with the model variables, necessary optimization objective function will be needed. In this paper, we introduce a hybrid learning task combo by following \cite{zhang2020graph}, which covers \textit{unsupervised node attribute reconstruction}, \textit{unsupervised graph structure recovery} and \textit{supervised node classification}.

\begin{itemize}

\item \textbf{Node Attribute Reconstruction}: Based on the learned node representations, via several fully connected layers (with necessary activation functions), we will be able to project the learned representation vectors to their raw features, i.e., the node raw attribute reconstruction. By minimizing the difference between nodes' original raw attributes versus the reconstructed ones, we will be able to learn the {\our} model. 

\item \textbf{Graph Structure Recovery}: Given any two nodes from the same graph, based on their learned representations, via either fully connected layers or simple similarity metrics, we will be able to project the node pair representation vectors to their corresponding link labels or similarity scores. Also by minimizing the differences between such learned link scores versus the graph link ground truth, {\our} can also be effectively learned.

\item \textbf{Node Classification}: In some cases, the nodes are also attached with labels denoting their categories or certain properties. Based on the nodes' representations, we can effectively project them to their desired labels via the fully connected layers (with necessary activation functions). By comparing such learned nodes' labels versus the node ground truth label vectors, we will be able to learn the {\our} model.

\end{itemize}

Considering that different graph datasets and different application tasks may require different learning parameter settings, instead of summing all the loss terms together to optimize, we introduce an iterative training mechanism for {\our} with the hybrid application tasks on all these multi-source graph inputs. To be more specific, for each graph source and in each iteration, we will train {\our} with a number epochs on the node classification task, and then on graph structure recovery task with several epochs, and finally a certain number epochs on the node classification task (the specific epoch numbers are different for different graph dataset, which will be introduced in Section~\ref{subsec:parameter_setting} in detail). Such an iterative training process will continue for several rounds on the graph source until there exist no dramatic changes as we shift between different learning tasks. After such a process, the {\our} model can be transferred and applied to certain graph sources for necessary fine-tuning.

\begin{figure}[t]
    \centering
    \begin{minipage}{.48\textwidth}
    	\includegraphics[width=\linewidth]{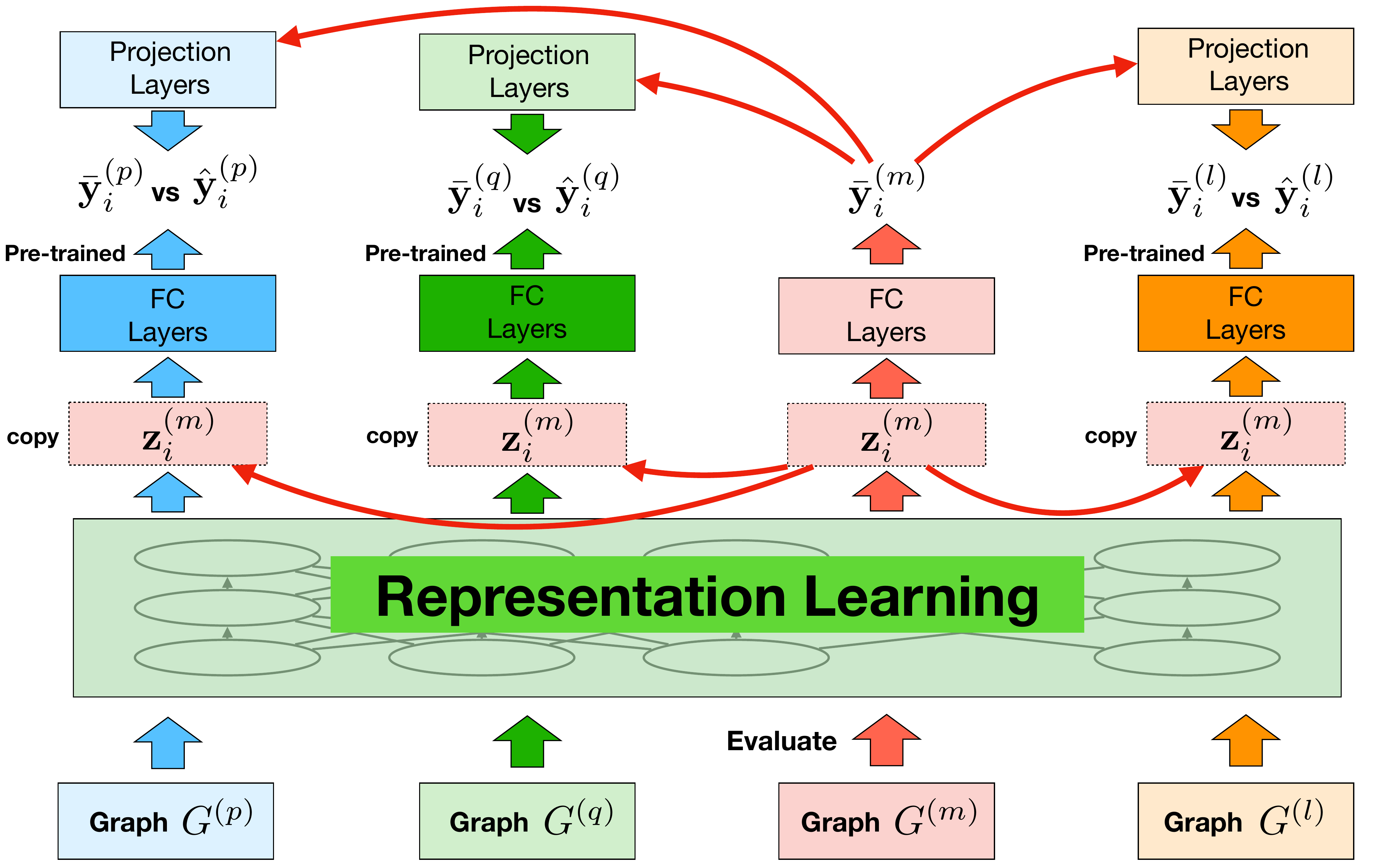}
     \end{minipage}%
        \caption{The reasoning process based on CCCM.}
    	\label{fig:reasoning1}
\end{figure}

\section{{\our} based Apocalypse Learning}\label{sec:analysis}

In this section, we will study a special and novel learning task, i.e., the \textit{apocalypse learning} problem, which aims at learning a classifier without using any labeled data. Such an open problem is intractable before, but the {\our} model actually provides us with the opportunity to explore it in this paper. In this part, we will introduce two different learning strategies, i.e., \textit{Cross-Source Classification Consistency Maximization} (CCCM) and \textit{Cross-Source Dynamic Routing} (CDR), to reason for the potential labels for the nodes in an input graph lacking supervision information, respectively.

\subsection{Reasoning Strategy \# 1: CCCM}


One approach proposed in this paper for the potential label reasoning for nodes in graph without supervision information is called the \textit{cross-source classification consistency maximization} (CCCM). Formally, as illustrated in Figure~\ref{fig:reasoning1}, let's take one of the target graph $G^{(m)}$ as an example, which contains no node labels, and we are studying the \textit{node classification} task based on it. Given the pre-trained {\our} model with several other graph datasets (containing supervised application functional components), via necessary fine-tuning with the other unsupervised learning tasks on $G^{(m)}$, e.g., \textit{node attribute reconstruction} and \textit{graph structure recovery}, we can still learn the representations of the nodes in the graph with {\our}, which can be representations as $\left\{\mb{z}_i^{(m)} \right\}_{v_i^{(m)} \in \mc{V}^{(m)}}$. Furthermore, for node $v_i^{(m)}$ with representation $\mb{z}_i^{(m)}$, via several fully connected layers, we can represent the node's label to be
\begin{equation}\label{equ:label_inference}
\bar{\mb{y}}_i^{(m)} = \mbox{softmax} \left( \mbox{FC}^{(m)} \left( \mb{z}_i^{(m)} \right) \right).
\end{equation}
According to the previous descriptions, with the input processing components for each datasets, the learned node representations from different graphs will lie in identical feature spaces actually. Based on such an intuition, via the learned {\our} models on the other graph datasets like $G^{(l)}$, given the node representation $\mb{z}_i^{(m)}$, we can also define their inferred labels by {\our} directly as $\left\{ \bar{\mb{y}}_i^{(l)}  \right\}_{l = 1 \land l \neq m}^{n}$, where 
\begin{equation}
\bar{\mb{y}}_i^{(l)} = \mbox{softmax} \left( \mbox{FC}^{(l)} \left( {\mb{z}}_i^{(m)} \right) \right).
\end{equation} 
Meanwhile, based on the inferred label vector $\bar{\mb{y}}_i^{(m)}$, we propose to project it to the other graph datasets via several FC layers, and the projected label vectors in the other datasets can be denoted as $\left\{ \hat{\mb{y}}_i^{(l)}  \right\}_{l = 1 \land l \neq m}^{n}$, where 
\begin{equation}
\hat{\mb{y}}_i^{(l)} = \mbox{softmax} \left( \mbox{FC}^{(m \to l)} \left( \bar{\mb{y}}_i^{(m)} \right) \right).
\end{equation}

In this paper, we assume that its learned class labels should carry consistent information across all these different graphs, since they are learned within the identical framework. Therefore, to learn the nodes label vectors in graph $G^{(m)}$ as well as the involved fully connected layers, we propose to minimize the below classification consistency loss term:
\begin{equation}
\min \sum_{v_i^{(m)} \in \mc{V}^{(m)}} \sum_{l = 1 \land l \neq m}^n \left\| \bar{\mb{y}}_i^{(l)} - \hat{\mb{y}}_i^{(l)} \right\|_2.
\end{equation}

\subsection{Reasoning Strategy \# 2: CDR}

The CCCM approach may need to learn several fully connected layers for the node label reasoning based on the classification result consistency assumption across graphs for common representation inputs. Here, in this part, we will introduce another reasoning approach based on the dynamic routing algorithm instead, which works very differently. Formally, for any node $v_i^{(m)}$ in graph $G^{(m)}$, we can denote its representation in $G^{(m)}$ as vectors $\mb{z}_i^{(m)}$. Furthermore, by feeding $\mb{z}_i^{(m)}$ as the input for classifiers in other graph sources, we can represent their learned label vectors as $\left\{ \bar{\mb{y}}_i^{(l)}  \right\}_{l = 1 \land l \neq m}^{n}$, respectively. The \textit{cross-source dynamic routing} (CDR) approach reasons nodes' labels in $G^{(m)}$ iteratively as follows:
\begin{equation}
\begin{cases}
\mb{c}_i &= \mbox{softmax} \left( \mb{b}_i \right), \\
\mb{u}_i^{(l \to m)} &= \mb{W}^{(l \to m)} \bar{\mb{y}}_i^{(l)} ,\\
\mb{s}_i &= \sum_{l} \mb{c}_i(l) \mb{u}_i^{(l \to m)}, \\
\mb{v}_i &= \frac{\left\| \mb{s}_i \right\|^2}{1+\left\| \mb{s}_i \right\|^2} \frac{\mb{s}_i}{ \left\| \mb{s}_i \right\|},\\
\mb{b}_i(l) &=  \mb{b}_i(l) + \mb{v}_i^\top \mb{u}_i^{(l \to m)}.
\end{cases}
\end{equation}
where $\mb{W}^{(l \to m)} \in \mathbbm{R}^{d_y^{(m)} \times d_y^{(l)}}$ denotes the label vector dimension adjustment variable between graphs $G^{(l)}$ and $G^{(m)}$. Formally, the vector $\mb{v}_i$ outputted by such a process will represent the reasoned label vector of node $v_i^{(m)}$. By minimizing its difference with the inferred label by {\our}, i.e., as defined in Equation~(\ref{equ:label_inference}), we will be able to represent the introduced reasoning loss function as follows:
\begin{equation}
\min \sum_{v_i^{(m)} \in \mc{V}^{(m)}} \left\| \bar{\mb{y}}_i^{(m)} - {\mb{v}}_i \right\|_2.
\end{equation}
More information about the experimental studies of these two \textit{apocalypse learning} oriented reasoning strategies will be provided in the following section in detail.

\section{Experiments}\label{sec:experiment}

To test the effectiveness of {\our} on graph representation learning, in this section, we will report some preliminary experimental results of {\our} that we obtain on three real-world benchmark graph datasets. More experimental results will be provided in the followup updated version of this paper as well.



\subsection{Dataset and Learning Settings}\label{subsec:parameter_setting}

The graph benchmark datasets used in the experiments include Cora, Citeseer and Pubmed \cite{YCS16}, which are used in most of the recent state-of-the-art graph neural network research works \cite{Kipf_Semi_CORR_16,Velickovic_Graph_ICLR_18,Li_Deeper_CORR_18,sun2019adagcn,DBLP:journals/corr/abs-1907-02586,Zhang2019GResNetGR}. For fair comparison, the experimental settings, e.g., train/validation/test set partition, will be identical as these existing research papers as well. Based on the input graph data, we will first pre-compute the node intimacy scores, based on which subgraph batches will be sampled subject to the subgraph size $k$ for each dataset. In addition, we will also pre-compute the node pairwise hop distance and WL node codes. In this paper, we aim to examine the transfer of the universal {\gbert} across different graph datasets based on the {\our} framework. Considering that different datasets will have different learning settings, instead of summing the loss functions of all the datasets, we propose to train {\our} with multiple graph datasets iteratively. To be more specific, the pre-training of {\our} will last for several iterations. In each iteration, we will train the corresponding components in {\our} with Cora, Citeseer and Pubmed sequentially subject to their unique parameter settings shown as follows. The default evaluation metric used in the experiments is Accuracy.

\noindent \textbf{Default Parameter Settings}: If not clearly specified, the results reported in this paper are based on the following parameter settings of {\our}: \textit{subgraph size}: $k=7$ (Cora), $k=5$ (Citeseer), $k=30$ (Pubmed); \textit{hidden size}: 32; \textit{attention head number}: 2; \textit{hidden layer number}: $D=2$; \textit{learning rate}: 0.01 (Cora) and 0.001 (Citeseer) and 0.001 (Pubmed); \textit{weight decay}: $5e^{-4}$; \textit{intermediate size}: 32; \textit{hidden dropout rate}: 0.5; \textit{attention dropout rate}: 0.3; \textit{graph residual term}: graph-raw; \textit{optimizer}: Adam; \textit{training epoch}: 150 (Cora), 500 (Pubmed), 2000 (Citeseer). For the universal {\gbert}, we evaluate the learning performance by changing its parameter $k$ with different values from $\{5, 7, 15, 30\}$ in the experiments, where $5$, $7$, $30$ are the optimal parameters for these three datasets, respectively, and value $15$ can balance among all the datasets.

\noindent \textbf{Experiment Organization}: We intend to use the experiments to answer several questions that readers may have in mind:
\begin{itemize}
\item \textbf{Q1}: Can {\our} still work well for isolated graph input?
\item \textbf{Q2}: Can {\our} be applicable to multiple graph inputs, which all have abundant training data actually?
\item \textbf{Q3}: How will the pre-trained {\our} perform when being transferred to target graphs lacking enough training data?
\item \textbf{Q4}: How is the learning performance of two reasoning strategies in {\our} on addressing the apocalypse learning task?
\end{itemize}
The following experiments will be designed to address these above above questions specifically.

%

\subsection{Isolated {\our} on Node Classification}

\begin{table}[t]
\caption{Learning performance of {\our} compared against existing baseline methods on node classification. The results of {\our} reported here denotes the best observed scores obtained on each dataset in the isolated mode.}\label{tab:isolated_learning}
\centering
\begin{tabular}{l c c c c }
\toprule
 \multirow{2}{*}{Methods}  & \multicolumn{3}{c}{Datasets (Accuracy)} \\
\cline{2-4}
\addlinespace[0.05cm]
& \textbf{Cora} & \textbf{Citeseer} & \textbf{Pubmed} \\
\hline
\addlinespace[0.05cm]
{LP (\cite{ZGL03}) } &0.680 &0.453 &0.630  \\
{ICA (\cite{LG03})} &0.751  &0.691  &0.739   \\
{ManiReg (\cite{BNS06})} &0.595  &0.601  &0.707   \\
{SemiEmb (\cite{WRC08})} &0.590  &0.596  &0.711  \\
\hline
\addlinespace[0.05cm]
{DeepWalk (\cite{PAS14})} &0.672  &0.432  &0.653   \\
{Planetoid (\cite{YCS16})} &0.757  &0.647  &0.772  \\
{MoNet (\cite{MBMRSB16})} &0.817  &-  &0.788  \\
\hline
\addlinespace[0.05cm]
{{\gcn} (\cite{Kipf_Semi_CORR_16})} &0.815  &0.703  &\textbf{0.790}   \\
{{\gat} (\cite{Velickovic_Graph_ICLR_18})} &\textbf{0.830}  &\textbf{0.725}  &\textbf{0.790}  \\
{{\loopy} (\cite{loopynet})} &{0.826}  &\textbf{0.716}  &\textbf{0.792}  \\
\hline
\addlinespace[0.05cm]
{\gbert} (\cite{zhang2020graph}) &\textbf{0.843}  &{0.712}  &\textbf{0.793}  \\
\hline
\addlinespace[0.05cm]
\multirow{2}{*}{{\our} (isolated)}
 &\textbf{0.841}	&\textbf{0.715}  &{0.789}  \\
\addlinespace[-0.05cm]
&($k=7$)&($k=5$)&($k=30$)\\
\bottomrule
\end{tabular}
\end{table}

Prior to showing the learning performance of {\our} across multiple graph datasets, we will first provide the learning results of {\our} on node classification based on each graph dataset in an isolated learning mode in Table~\ref{tab:isolated_learning}. The isolated version of {\our} is very similar to {\gbert} studied in \cite{zhang2020graph} actually, except that {\our} will have two more graph-transformer layers (i.e., the input processing component for each dataset) besides the shared universal {\gbert} component. To make the comparison more complete, in addition to {\gbert} \cite{zhang2020graph}, we also provide the learning results of several classic graph classification methods, e.g., LP \cite{ZGL03}, ICA \cite{LG03}, ManiReg \cite{BNS06}, SemiEmb \cite{WRC08}, recent graph embedding methods, DeepWalk \cite{PAS14}, Planetoid \cite{YCS16}, MoNet \cite{MBMRSB16}, and the latest graph representation learning approaches, e.g., {\gcn} \cite{Kipf_Semi_CORR_16}, {\gat} \cite{Velickovic_Graph_ICLR_18}, {\loopy} \cite{loopynet}. According to the results, the scores achieved by {\our} are very close to those of {\gbert}, which are much higher than the scores obtained by the other baseline methods.

\subsection{Results of {\our} on Mixed Graph Input}

\begin{table}[t]
\caption{Learning performance of {\our} with a mixed pre-training for node classification on multiple input graph datasets. Parameter $k$ denotes the input portal size of the universal {\gbert} component.}\label{tab:mixed_learning}
\centering
\begin{tabular}{|c|c|c|c|c|}
\hline
\multicolumn{2}{|c}{{Input Graphs} \& $k$ } & \multicolumn{3}{|c|}{Datasets (Accuracy)} \\
\hline
Graphs & $k$ & {\textbf{Cora}} & {\textbf{Citeseer}} & {\textbf{Pubmed}} \\
\hline 
\hline 
\multirow{4}{*}{Cora \& Citeseer}
&5 &{0.834}   &{0.707}  &$-$   \\
\cline{2-5}
&7 &\textbf{0.835}   &\textbf{0.717}  &$-$   \\
\cline{2-5}
&15 &0.828   &{0.702}  &$-$   \\
\cline{2-5}
&30&{0.822}   &{0.698}  &$-$  \\
\hline
\hline 
\multirow{4}{*}{Cora \& Pubmed}
&5 &\textbf{0.832}   &$-$	&{0.772}   \\
\cline{2-5}
&7 &0.828   &$-$	&{0.766}     \\
\cline{2-5}
&15 &0.829   &$-$	&{0.782}   \\
\cline{2-5}
&30	&{0.816}	&$-$	&\textbf{0.791}	\\
\hline
\hline 
\multirow{4}{*}{Citeseer \& Pubmed}
&5 &$-$   &\textbf{0.705}   &{0.772}  \\
\cline{2-5}
&7 &$-$   &0.702   &{0.773}  \\
\cline{2-5}
&15 &$-$   &0.683   &{0.787}  \\
\cline{2-5}
&30&$-$	&{0.675}   &\textbf{0.782}  \\
\hline
\end{tabular}
\end{table}

\begin{table*}[t]
\caption{Learning performance of {\our} with model transfer. The source graphs are for {\our} pre-training, and the target graph are used for {\our} evaluation with necessary fine-tuning. We focus on studying the effectiveness of {\our} transfer to the target graph with sparse training data, where the training data sampling ratio denotes the percentage of training data used for model fine-tuning. For comparison, we also illustrate the learning performance of {\our} without pre-training at all in the table.}\label{tab:transfer}
\centering
\setlength{\tabcolsep}{8pt}
\begin{tabular}{|c|c|c|c|c|c|c|c|c|c|c|c|c|}
\hline
\multicolumn{3}{|c}{\textbf{Source Graph(s)} \& \textbf{Target Graph} \& $k$ } & \multicolumn{10}{|c|}{\textbf{Training Data Sampling Ratio (Accuracy)}} \\
\hline
\textbf{Source(s)} & \textbf{Target} & $k$ & \textbf{{5\%}} & \textbf{{10\%}} & \textbf{{15\%}} & \textbf{{20\%}} & \textbf{{25\%}} & \textbf{{30\%}} & \textbf{{35\%}} & \textbf{{40\%}} & \textbf{{45\%}} & \textbf{{50\%}} \\
\hline 
\hline 
\multirow{2}{*}{Cora}
&Citeseer &15 &0.418	&0.569	&0.541	&0.546	&0.557	&0.600	&0.593	&0.607	&0.623	&0.661   \\
\cline{2-13}
&Pubmed &15 &0.530	&0.649	&0.669	&0.692	&0.692	&0.687	&0.692	&0.697	&0.710	&0.743   \\
\hline 
\hline 
\multirow{2}{*}{Citeseer}
&Cora &15 &0.262	&0.420	&0.546	&0.619	&0.684	&0.662	&0.706	&0.727	&0.729	&0.748   \\
\cline{2-13}
&Pubmed &15 &0.524	&0.692	&0.697	&0.682	&0.723	&0.717	&0.736	&0.744	&0.740	&0.741   \\
\hline 
\hline 
\multirow{2}{*}{Pubmed}
&Cora &15 &0.317	&0.405	&0.551	&0.559	&0.740	&0.753	&0.747	&0.759	&0.804	&0.805   \\
\cline{2-13}
&Citeseer &15 &0.362	&0.583	&0.553	&0.553	&0.643	&0.626	&0.624	&0.620	&0.616	&0.667   \\
\hline
\hline 
\multirow{1}{*}{Cora \& Citeseer}
&Pubmed &15 &0.501	&0.662	&0.643	&0.658	&0.655	&0.667	&0.664	&0.670	&0.659	&0.672   \\
\hline
\multirow{1}{*}{Cora \& Pubmed}
&Citeseer &15 &0.368	&0.571	&0.584	&0.573	&0.572	&0.586	&0.584	&0.590	&0.595	&0.698   \\
\hline
\multirow{1}{*}{Citeseer \& Pubmed}
&Cora &15 &0.300	&0.456	&0.544	&0.662	&0.746	&0.765	&0.778	&0.769	&0.787	&0.784   \\
\hline
\hline 
\multirow{3}{*}{None (No Pre-train)}
&Cora &7 &0.299	&0.404	&0.480	&0.574	&0.701	&0.688	&0.706	&0.768	&0.777	&0.794   \\
\cline{2-13}
&Citeseer &5 &0.341	&0.567	&0.541	&0.553	&0.558	&0.580	&0.583	&0.582	&0.598	&0.637   \\
\cline{2-13}
&Pubmed &30 &0.485	&0.630	&0.638	&0.617	&0.604	&0.608	&0.608	&0.572	&0.599	&0.641   \\
\hline
\end{tabular}
\end{table*}

\begin{table}[t]
\caption{Reasoning performance of {\our} with different strategies for apocalypse learning (``Random'': random guess).}\label{tab:reasoning}
\centering
\begin{tabular}{|c|c|c|c|c|}
\hline
\multicolumn{2}{|c}{\textbf{Source \& {Target Graph(s)}} } & \multicolumn{3}{|c|}{\textbf{Reaning Strategies}}\\
\hline
\textbf{Source(s)} & \textbf{Target} & \textbf{CCCM} & \textbf{CDR} & \textbf{Random}\\
\hline 
\hline 
\multirow{2}{*}{Cora}
&Citeseer &0.280   &\textbf{0.312}	&0.167	\\
\cline{2-5}
&Pubmed &0.551   &\textbf{0.544}   &0.333	\\
\hline 
\multirow{2}{*}{Citeseer}
&Cora &0.323   &\textbf{0.358}  &0.143	\\
\cline{2-5}
&Pubmed &0.505   &\textbf{0.515}  &0.333	\\
\hline 
\multirow{2}{*}{Pubmed}
&Cora &\textbf{0.342}   &{0.304}  &0.143	\\
\cline{2-5}
&Citeseer &0.323   &\textbf{0.331}   &0.167	\\
\hline 
{Cora \& Citeseer}
&Pubmed &0.516  &\textbf{0.519}  &0.333	\\
\hline
{Cora \& Pubmed}
&Citeseer &0.318   &\textbf{0.332}   &0.167	\\
\hline
{Citeseer \& Pubmed}
&Cora &\textbf{0.327}   &{0.319}  &0.143	\\
\hline
\end{tabular}
\end{table}

In Table~\ref{tab:mixed_learning}, we provide the learning results of {\our} learned with multiple graph inputs. To be more specific, given the input graphs, we will pre-train {\our} with the hybrid application tasks on these graph datasets. Such pre-trained {\our} model will be further fine-tuned on each graph for the node classification task. For each graph, the parameter $k$ of its input pre-processing component is assigned with the default parameter as introduced before. Meanwhile, for the universal {\gbert} involved in {\our}, we change its input size parameter $k$ with values in $\{5, 7, 15, 30\}$, where $5$, $7$ and $30$ are the optimal parameter $k$ for Citeseer, Cora and Pubmed, respectively, and value $15$ balances among these optimal parameters.

According to the results, we observe that training {\our} concurrently with multiple input graphs and hybrid application tasks will have some minor impacts on its performance on the node classification task. In some cases, compared with Table~\ref{tab:isolated_learning}, there are some drops of the scores, e.g., {\our} on Cora. Meanwhile, in some other cases, the learning performance of {\our} can also be very good, which are highlighted in the table. What's more, parameter $k$ of the universal {\gbert} model does have an impact on the performance of {\our}, where Cora and Citeseer favor small $k$, whereas Pubmed prefers larger $k$ instead. To achieve the balanced performance, we will set $k=15$ for the following studies on {\our} transfer across different graph datasets.

\subsection{Transfer of {\our} to Sparsely Labeled Graph}

In Table~\ref{tab:transfer}, we provide the learning results of {\our} on graphs with sparse labels. To be more specific, we will pre-train {\our} on the source graphs with the hybrid application tasks and transfer the pre-trained model to the target graph(s) for evaluation. Since we focus on the graphs with sparse labels, a small portion of the labeled data are sampled from the target graph for model fine-tuning, where the sampling ratio changes with value in $\{$5\%$, $10\%$, \cdots, $50\%$\}$. Meanwhile, for comparison completeness, we also provide the results of {\our} without pre-training in the table, where the parameter $k$ of the universal component is assigned with the optimal values favored by the graph datasets. According to the results, in most of the cases, {\our} with pre-training can out-perform that without pre-training consistently.

\subsection{Reasoning of {\our} for Apocalypse Learning}

In Table~\ref{tab:reasoning}, we provide the learning results of {\our} based on the apocalypse learning settings, where the target graph has no labeled data at all. All the existing graph neural networks will fail to work in such a learning setting. To enable {\our} can work to address the node classification problem on the target graph, we pre-train {\our} on the source graphs to learn the universal {\gbert} component shared across graphs. Furthermore, such pre-trained {\our} will be further fine-tuned on the target graph with the unsupervised learning tasks, i.e., \textit{node attribute reconstruction} and \textit{graph recovery}, so as to learn the input component for the target graph in {\our}. Based on the CCCM and CDR reasoning strategies, {\our} will still be able to reason for the potential labels for the nodes in the target graph. For comparison, we also provide the results of random guess in the table, and the scores achieved by {\our} with these two reasoning strategies are both much higher than random guess.

\section{Related Work}\label{sec:related_work}

Several interesting research topics are related to this paper, which include \textit{graph neural network} and \textit{{\bert}}.

\noindent \textbf{Graph Neural Network}: In addition to the graph convolutional neural network \cite{Kipf_Semi_CORR_16} and its derived variants \cite{Velickovic_Graph_ICLR_18,sun2019adagcn,DBLP:journals/corr/abs-1907-02586}, many great research works on graph neural networks have been witnessed in recent years for graph representation learning \cite{SPIGCN,Zhang2018AnED,Ivanov_Anonymous_18,xinyi2018capsule}. Many existing graph neural network models will suffer from performance problems with deep architectures. In \cite{Zhang2019GResNetGR,Li_Deeper_CORR_18,sun2019adagcn,Huang_Inductive_19}, the authors explore to build deep graph neural networks with residual learning, dilated convolutions, and recurrent network, respectively. In \cite{zhang2020graph}, the authors introduce a new type of graph neural network based on graph transformer and BERT, i.e., the {\gbert} model. Different from the node representation learning \cite{Kipf_Semi_CORR_16,Velickovic_Graph_ICLR_18}, GNNs proposed for the graph representation learning aim at learning the representation for the entire graph instead \cite{Narayanan_Graph_17}. To handle the graph node permutation invariant challenge, solutions based various techniques, e.g., attention \cite{Chen_Dual_19,Meltzer_Permutation_19}, pooling \cite{Meltzer_Permutation_19,ranjan2019asap,Jiang_Gaussian_18}, capsule net \cite{Mallea_Capsule_19}, Weisfeiler-Lehman kernel \cite{NIPS2016_6166} and sub-graph pattern learning and matching \cite{Meng_Isomorphic_NIPS_19}, have been proposed. To apply {\gbert} on graph instance modeling and handle diverse graph instance sizes, \cite{zhang2020segmented} proposes several different graph instance size unification approaches. 


\noindent \textbf{{\bert}}: {\transformer} \cite{Vaswani_Attention_17} and {\bert} \cite{Bert} based models have almost dominated NLP and related research areas in recent years due to their great representation learning power. Prior to that, the main-stream sequence transduction models in NLP are mostly based on complex recurrent \cite{Hochreiter_Long_Neural_97,DBLP:journals/corr/ChungGCB14} or convolutional neural networks \cite{kim-2014-convolutional}. However, as introduced in \cite{Vaswani_Attention_17}, the inherently sequential nature precludes parallelization within training examples. To address such a problem, a brand new representation learning model solely based on attention mechanisms, i.e., the {\transformer}, is introduced in \cite{Vaswani_Attention_17}, which dispense with recurrence and convolutions entirely. Based on {\transformer}, \cite{{Bert}} further introduces {\bert} for deep language understanding, which obtains new state-of-the-art results on eleven natural language processing tasks. By extending {\transformer} and {\bert}, many new {\bert} based models, e.g., T5 \cite{raffel2019exploring}, ERNIE \cite{Sun_ERNIE} and RoBERTa \cite{Liu_RoBERTa}, can even out-perform the human beings on almost all NLP benchmark datasets. 

 \section{Conclusion}\label{sec:conclusion}

In this paper, we have studied the graph-to-graph transfer of a universal {\gbert} for graph representation learning across different graph datasets. To address the problem, we introduce a new learning model named {\our}, whose pluggable architecture containing several key parts, i.e., (1) pluggable input dataset-wise components, (2) input size unification interlayer, (3) the universal Graph-Bert model shared across graphs, (4) representation fusion interlayer; (5) pluggable task-wise output components for each dataset, and (6) reasoning component for \textit{apocalypse learning}. Furthermore, based on the {\our} model, we also investigate a special and novel learning task, i.e., the \textit{apocalypse learning} problem, which aims at learning a classifier without using any labeled data. Two different reasoning strategies, i.e., CCCM and CDR, are proposed to reason for the potential labels for the nodes. To test the effectiveness of {\our}, some preliminary experiments have been done on real-world graph datasets and the results also demonstrate the effectiveness of both {\our} and these two proposed reasoning strategies.

\newpage
{
\bibliographystyle{abbrv}
\bibliography{reference}
}


\end{document}